\documentclass[sigconf]{acmart}
\AtBeginDocument{%
  }

\setcopyright{acmlicensed}
\copyrightyear{2025}
\acmYear{2025}
\acmDOI{XXXXXXX.XXXXXXX}
\acmConference[ACM MM '25]{Proceedings of the 33rd ACM International Conference on Multimedia}{October 27--31,
  2025}{Dublin, Ireland}
\acmISBN{978-1-4503-XXXX-X/2018/06}

\acmSubmissionID{6949}

\usepackage{tabularray}

\begin{document}

\title{GaussVideoDreamer: 3D Scene Generation with Video Diffusion and Inconsistency-Aware Gaussian Splatting}

\author{Junlin Hao}
\affiliation{%
  \institution{Peking University}
  \city{Beijing}
  \country{China}
}
\email{junlin.hao@stu.pku.edu.cn}

\author{Peiheng Wang}
\affiliation{%
  \institution{Peking University}
  \city{Beijing}
  \country{China}
}
\email{peiheng.wang@pku.edu.cn}

\author{Haoyang Wang}
\affiliation{%
  \institution{Peking University}
  \city{Beijing}
  \country{China}
}
\email{haoyang.wang@stu.pku.edu.cn}

\author{Xinggong Zhang}
\affiliation{%
  \institution{Peking University}
  \city{Beijing}
  \country{China}
}
\email{zhangxg@pku.edu.cn}

\author{Zongming Guo}
\affiliation{%
  \institution{Peking University}
  \city{Beijing}
  \country{China}
}
\email{guozongming@pku.edu.cn}







\begin{abstract}
Single-image 3D scene reconstruction presents significant challenges due to its inherently ill-posed nature and limited input constraints. Recent advances have explored two promising directions: \textbf{multiview generative models} that train on 3D consistent datasets but struggle with out-of-distribution generalization, and \textbf{3D scene inpainting and completion} frameworks that suffer from cross-view inconsistency and suboptimal error handling, as they depend exclusively on depth data or 3D smoothness, which ultimately degrades output quality and computational performance. Building upon these approaches, we present \textbf{GaussVideoDreamer}, which advances generative multimedia approaches by bridging the gap between image, video, and 3D generation, integrating their strengths through two key innovations: (1) A progressive video inpainting strategy that harnesses temporal coherence for improved multiview consistency and faster convergence. (2) A 3D Gaussian Splatting consistency mask to guide the video diffusion with 3D consistent multiview evidence. Our pipeline combines three core components: a geometry-aware initialization protocol, Inconsistency-Aware Gaussian Splatting, and a progressive video inpainting strategy. Experimental results demonstrate that our approach achieves 32\% higher LLaVA-IQA scores and at least 2x speedup compared to existing methods while maintaining robust performance across diverse scenes.
\end{abstract}

\ccsdesc[100]{Computing methodologies~Reconstruction}
\ccsdesc[300]{Computing methodologies~Generative and developmental approaches}
\ccsdesc[300]{Computing methodologies~Graphics systems and interfaces}
\keywords{Diffusion Model, 3D Reconstruction, Gaussian Splating, Novel View Synthesis, Generative Model}
\begin{teaserfigure}
  \includegraphics[width=\textwidth]{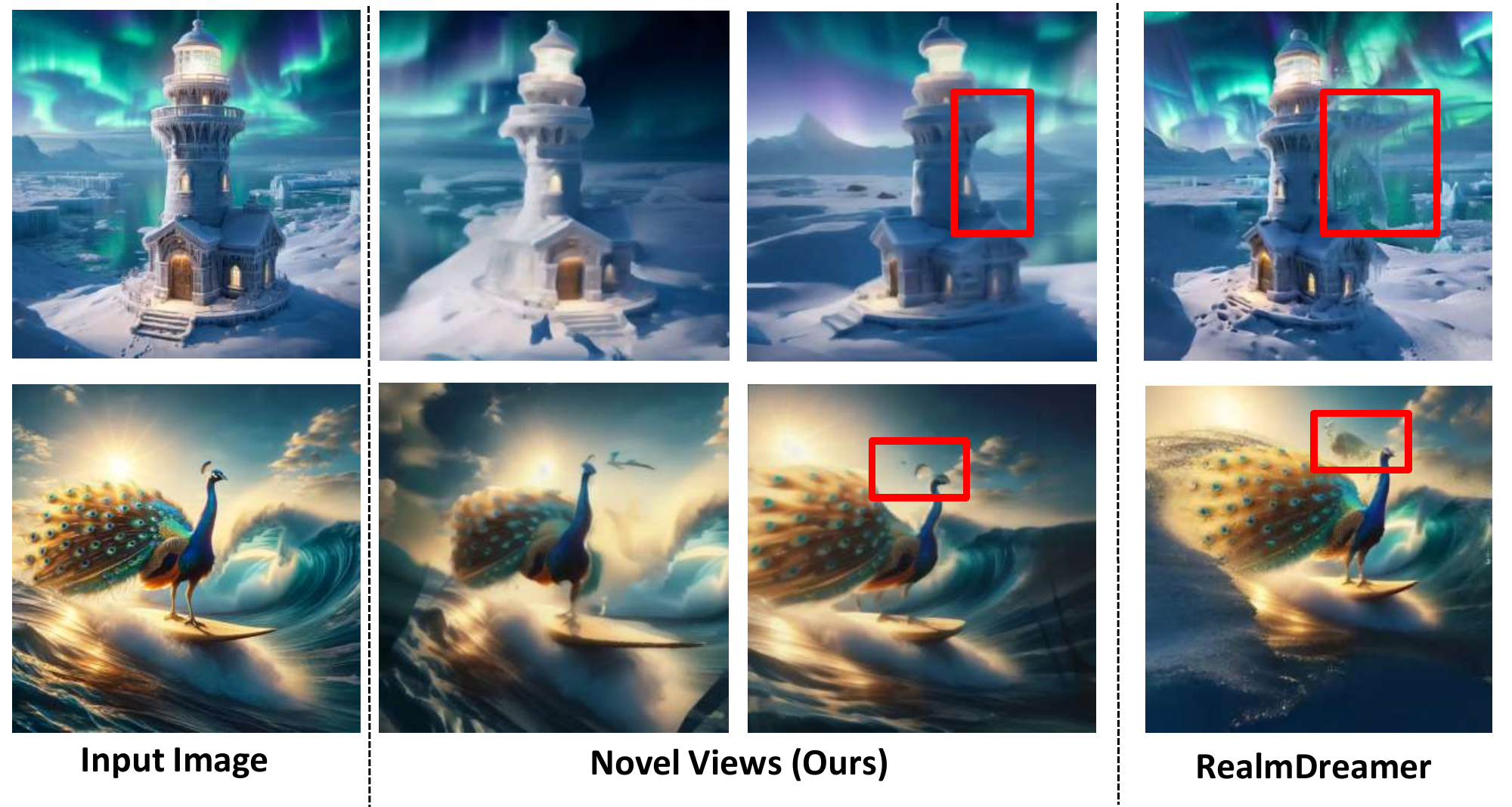}
  \caption{Given a single input image, GaussVideoDreamer generates high-quality 3D scenes with faster processing and improved scene context understanding compared to our baselines  \cite{shriram2024realmdreamer}, especially in the occluded and background regions.}
  \Description{Teaser.}
  \label{fig:teaser}
\end{teaserfigure}


\maketitle

\section{Introduction}
3D scene reconstruction has long been a fundamental problem in computer vision and graphics, with significant application potential in virtual reality, autonomous driving, and robotics. Traditional 3D reconstruction algorithms \cite{dust3r_cvpr24, kerbl3Dgaussians, mildenhall2020nerf, schoenberger2016sfm, schoenberger2016mvs} require dense multi-view images or specialized equipment to capture scene information like RGB-D data. However, in most practical scenarios, only limited or even just a single image is available. This poses a challenging and ill-posed problem for achieving high-quality 3D reconstruction under constrained data conditions.

With the rapid development of large pre-trained models such as Large Diffusion Models (LDM) \cite{rombach2022high, saharia2022photorealistic, podell2023sdxl}, this long-standing challenge has been primarily tackled via two generative paradigms. A large number of works \cite{zeronvs, gao2024cat3d, yu2024viewcrafter, charatan2024pixelsplat3dgaussiansplats, szymanowicz2025bolt3dgenerating3dscenes} propose \textbf{multiview generative models}, which incorporate 3D consistency constraints within the model and train the model on 3D consistent data \cite{ma2025itgotitlearning, reizenstein2021commonobjects3dlargescale} to learn generalized 3D reasoning capabilities.
These methods can obtain novel view images or 3D representations in a single process, eliminating the need for iterative scene optimization. Meanwhile, another line of research \cite{shriram2024realmdreamer, wu2023reconfusion, yu2025wonderworld, perf2023} focuses on leveraging diffusion models for \textbf{3D scene inpainting and completion}, where the model iteratively generates novel view information to refine and inpaint the missing regions in the scene. These works explicitly decouple 3D consistency constraints from the diffusion model by introducing an external 3D field, and this field provides geometric supervision to the diffusion process while the diffusion model reciprocally completes the 3D scene through iterative optimization.

Although remarkable progress has been achieved, current approaches still face several domain-specific challenges that hinder their practical applications.

\begin{figure}[h]
    \centering
    \includegraphics[width=\linewidth]{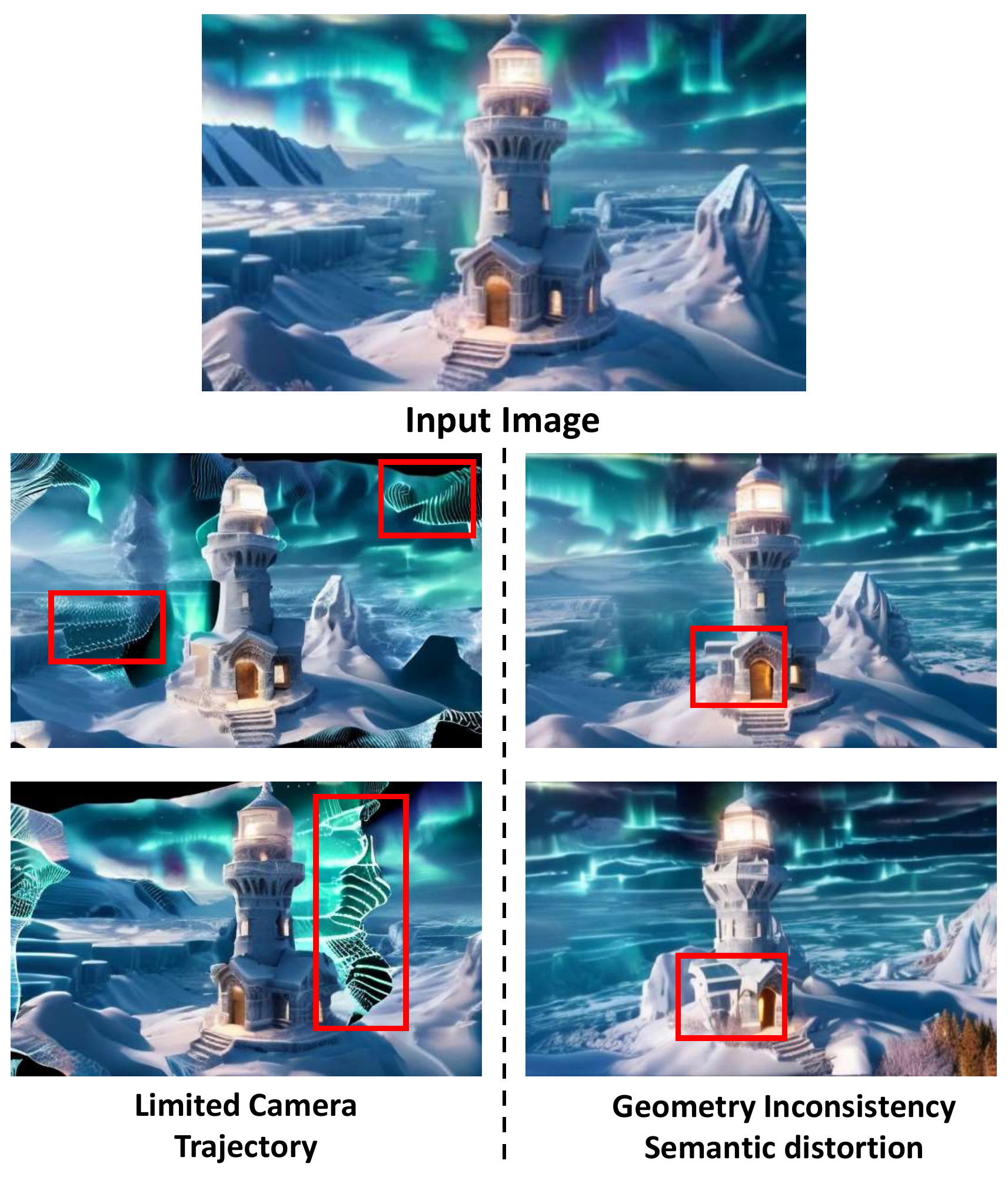}
    \Description{Fine-tuned diffusion's generalization capability}
    \caption{Multiview generative models exhibit limited generalization capability, manifesting as geometric inconsistencies and semantic distortions in synthesized views.}
    \label{fig:mvdiff_error}
\end{figure}

In the context of multiview generative models, two primary weaknesses persist: \textbf{(1)} Limited generalization capability: While large generative models acquire powerful priors through extensive training, existing multiview generative methods exhibit poor performance on out-of-distribution data \cite{gao2024cat3d} due to scarce 3D-consistent datasets and inefficient 3D-to-2D encoding mechanisms. \textbf{(2)} Information-scarce degradation: Under limited visual information conditions, particularly during large viewpoint variations, multiview generative models exhibit quality degradation in occluded regions, resulting in fractured geometry and semantic distortions. This reveals their limited generalization capability and inherent dependency on adequate input cues, shown in Fig.\ref{fig:mvdiff_error}.

When shifting focus to 3D scene inpainting and completion methods, distinct weaknesses emerge: \textbf{(1)} Single-view conditioning bias: Most existing works \cite{shriram2024realmdreamer, yu2025wonderworld} employ image diffusion models that condition scene context solely on individual viewpoints and are unaware of other novel views. This inherently causes inconsistent generation across different views, compromising scene quality and increasing required optimization iterations. \textbf{(2)} Crude inconsistency handling: Existing methods employ oversimplified inconsistency processing, typically using either occlusion detection from depth \cite{perf2023} or 3D field smoothness \cite{shriram2024realmdreamer, wu2023reconfusion}. They fail to comprehensively analyze visual inconsistencies across images, which leads to substantial information underutilization. As a result of these constraints, the optimization process suffers from \textbf{severely slow convergence}, needing multiple hours to refine just one scene.

\begin{figure*}[h!]
    \centering
    \includegraphics[width=\linewidth]{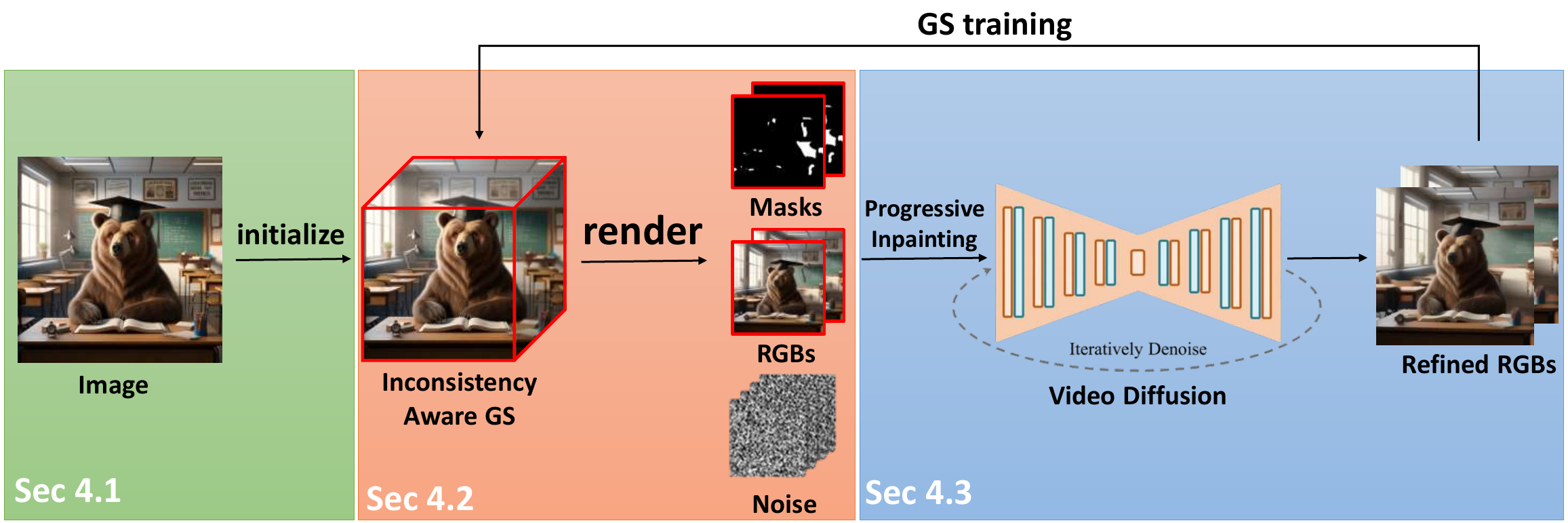}
    \Description{Our pipeline}
    \caption{Overview of our pipeline. Our method first initializes a coarse video and inconsistency-aware GS (IA-GS) from a single input image (Sec. \ref{sec:init}). At periodic optimization intervals, we render all viewpoint images and their corresponding inconsistency prediction masks from the IA-GS representation (Sec. \ref{sec:iags}). These masks and rendered images then guide a video diffusion model to perform progressive inpainting, editing regions based on their inconsistency levels (Sec. \ref{sec:refine}). The refined video sequence subsequently optimizes our IA-GS module and gradually generates better novel view images.}
    \label{fig:pipeline}
\end{figure*}

To address the aforementioned limitations in existing works, we pursue the technical trajectory of 3D scene inpainting and completion methods and propose GaussVideoDreamer (Fig. \ref{fig:pipeline}). Our key insight is that (1) video diffusion models inherently leverage richer scene context through temporal coherence, and (2) regions with higher reconstruction loss indicate geometric inconsistency, serving as effective inpainting guidance. While integrating video diffusion provides valuable temporal coherence priors, its direct application faces challenges, such as the inherent lack of 3D geometric awareness, which causes distortion and inconsistency in synthesized novel views. To solve these challenges, our method combines three key components: First, a geometry-aware initialization protocol establishes a robust coarse 3D scene to promote both quality and efficiency. Second, Inconsistency-Aware Gaussian Splatting (IA-GS) jointly optimizes scene representation with a learnable error predictor, which serves as a geometric consistency checker and adaptively guides the diffusion model's inpainting process. Third, a progressive inpainting strategy that gradually incorporates reliable multiview evidence to enhance the video diffusion model's generation quality. Extensive quantitative evaluations demonstrate that our approach achieves \textbf{32\% quality improvement} (LLaVA-IQA) and \textbf{at least 2× faster generation speed} compared to existing methods. Concretely, our contributions are the following:

1. Temporal-to-View Coherence: Replacing image diffusion with video diffusion to harness its temporal coherence as an inductive bias for multiview consistency and holistic scene understanding.

2. Advanced Inconsistency Handling: Our Inconsistency-Aware GS jointly optimizes scene representation and a learned predictor that adaptively guides video diffusion with progressive inpainting.

3. Our single-image-to-3D pipeline outperforms existing methods on out-of-distribution data, achieving 
 higher reconstruction quality while significantly reducing processing time, as validated by several quantitative metrics.

\section{Related Works}

\subsection{Multiview Generative Model}

Recent breakthroughs in large-scale transformer-based architectures and diffusion models \cite{rombach2022high, saharia2022photorealistic, podell2023sdxl, peebles2023scalablediffusionmodelstransformers} have shown remarkable adaptability for novel view synthesis and 3D generation from a single image, with contemporary research pursuing three principal methodological strands. The first line of research \cite{liu2023zero1to3, liu2023syncdreamer, gao2024cat3d, zeronvs, tewari2023diffusionforwardmodelssolving, Nguyen2025PointmapConditionedDF, hu2024mvdfusionsingleview3ddepthconsistent} enhances image diffusion models with explicit 3D awareness to be novel view generators, as exemplified by Zero-1-to-3 \cite{liu2023zero1to3}, which develops camera pose conditioning trained on synthetic datasets,  and SyncDreamer \cite{liu2023syncdreamer} which achieves synchronized multiview consistent generation through learnable 3D geometric constraints. Alternative approaches \cite{hong2024lrmlargereconstructionmodel, charatan2024pixelsplat3dgaussiansplats, szymanowicz2025bolt3dgenerating3dscenes, wewer2024latentsplatautoencodingvariationalgaussians, zhang2024gslrmlargereconstructionmodel, szymanowicz2024flash3dfeedforwardgeneralisable3d, Chen_2024_mvsplat} focus on end-to-end 3D representation learning and generation, demonstrated by LRM's \cite{hong2024lrmlargereconstructionmodel} pioneer transformer-based triplane-NeRF \cite{chan2022efficientgeometryaware3dgenerative} generation from single images and PixelSplat's \cite{charatan2024pixelsplat3dgaussiansplats} innovation through per-pixel Gaussian Splat parameter prediction. Concurrently, a third emerging paradigm \cite{wu2024cat4dcreate4dmultiview, liu2024reconxreconstructscenesparse, sun2024dimensionx, yu2024viewcrafter, wang2024motionctrlunifiedflexiblemotion, he2024cameractrl, guo2023sparsectrl} adapts video diffusion architectures to utilize inter-frame consistency with additional conditions like RGB \cite{xing2023dynamicrafteranimatingopendomainimages, blattmann2023stablevideodiffusionscaling, xing2024tooncraftergenerativecartooninterpolation, yu2024viewcrafter, müller2024multidiffconsistentnovelview}, depth \cite{xing2023makeyourvideocustomizedvideogeneration, esser2023structurecontentguidedvideosynthesis}, and semantic maps \cite{peruzzo2024vaseobjectcentricappearanceshape}, where MotionCtrl \cite{wang2024motionctrlunifiedflexiblemotion} achieves viewpoint control via camera extrinsic conditioning, and ViewCrafter \cite{yu2024viewcrafter} leverages eplicit pointcloud-render inpainting for precise camera control in video generation. Although enabling efficient generation in a single process, these approaches face persistent challenges, including  scarcity in 3D consistent training data and information degradation in 3D-to-2D encoding \cite{szymanowicz2025bolt3dgenerating3dscenes, ma2025itgotitlearning}, which harms their generalization capabilities \cite{gao2024cat3d} and demands prohibitive computational costs during large model training.

\subsection{3D Scene Inpainting and Completion}

The advent of DreamFusion  \cite{poole2022dreamfusiontextto3dusing2d} established another predominant 3D generation paradigm, introducing Score Distillation Sampling (SDS) to leverage 2D diffusion priors and optimize 3D scenes iteratively through single-step sampling from noisy images rendered at various viewpoints. Subsequent advances \cite{Deng2022NeRDiSN, Gu2023NerfDiffSV, melaskyriazi2023realfusion360degreconstructionobject, tang2023makeit3dhighfidelity3dcreation} have demonstrated remarkable success in object-level reconstruction.  Recent efforts \cite{wu2023reconfusion, chung2023luciddreamer, yu2025wonderworld, wang2024vistadreamsamplingmultiviewconsistent, chen2025flexworldprogressivelyexpanding3d, wang2023prolificdreamerhighfidelitydiversetextto3d, yu2024wonderjourney, zhang2023text2nerf} have extended this framework to scene-level inpainting, though the increased complexity introduces significant challenges. Unlike object-centric cases, scene reconstruction requires handling extensive content with strict geometric coherence,  leading most methods to adopt a coarse-to-refine strategy: first constructing a coarse 3D representation with depth estimation \cite{depthanything, bhat2023zoedepthzeroshottransfercombining}, then completing it via diffusion guidance and inpainting loss. However, two fundamental limitations persist: image-guided diffusion models inherently lack holistic scene understanding, resulting in view-inconsistent inpainting, and the absence of explicit inconsistency detection mechanisms \cite{sabour2024robustnerfignoringdistractorsrobust} forces SDS to inefficiently address geometric errors, resulting in unstable optimization, manifesting as blur artifacts and excessive compute demands.

\begin{figure*}[h]
    \centering
    \includegraphics[width=\linewidth]{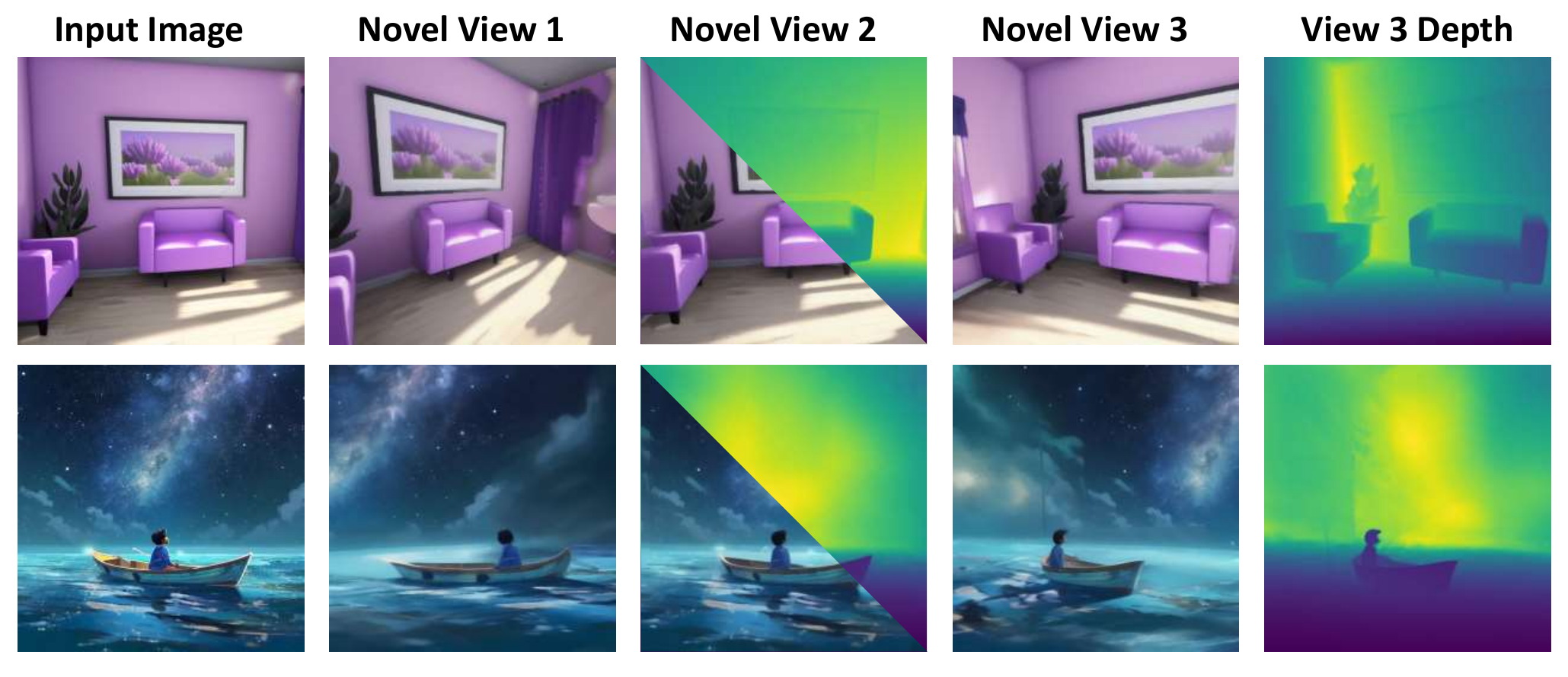}
    \Description{Qualitative Results}
    \caption{Qualitative Results. Left: Input reference image. Middle: Novel view renderings and RGB-depth split from our 3DGS. Right: Depth map visualization. The results demonstrate that our method can generate coherent color and consistent geometry across both indoor and outdoor scenes.}
    \label{fig:quali_res}
\end{figure*}

\section{Preliminary}

\paragraph{\textbf{Notation Convention}} 

We maintain distinct notations for diffusion and Gaussian Splats optimization processes throughout our formulations. All superscript $(\mathrm{t})$ denote GS iterations, while subscript $t$ indicates diffusion timesteps.

\subsection{Distractor-Free 3D Gaussian Splatting}

We build our pipeline and technique on top of 3D Gaussian Splating (3DGS) \cite{kerbl3Dgaussians}, a competitive alternative to NeRF \cite{mildenhall2020nerf}, with real-time rendering speeds and memory-efficient training. It represents the 3D scene as collections of anisotropic 3D Gaussians (splats) $\mathcal{G}$. Each splat's geometry is represented by a mean $\mu \in R^3$, scale vector $s \in \mathbb{R}^3$, and quaternion $q \in \mathbb{R}^4$, so that the covariance, i.e. the shape of the splat is given by $\Sigma=RSS^TR^T$, where $S = \text{Diag}(s)$ and $R$ is the rotation matrix computed from $q$. Each splat also has a corresponding opacity $\sigma \in \mathbb{R}$ and color $c \in \mathbb{R}^3$.

 Given posed images $\{I_i\}^N_{i=1}$, $I_i \in \mathbb {R}^{H \times W}$, 3DGS optimizes these splats through volumetric rendering. Once splat positions and covariances in screen spaces are computed, the pixel color $C(\mathbf{p})$ at coordinate $\mathbf{p}$ is computed via alpha-blending:

\begin{displaymath}
C(\mathbf{p})=\sum_{i \in \mathcal{N}} \alpha_i c_i \prod_{j=1}^{i-1}\left(1-\alpha_j\right), \quad \alpha_i=\sigma_i e^{-\frac{1}{2}\left(\mathbf{p}-\mu_i\right)^T \sum^{-1}\left(\mathbf{p}-\mu_i\right)},
\end{displaymath}

where $\mathcal{N}$ denotes splats overlapping p, and $\alpha_i$ implements a Gaussian-weighted opacity. While 3DGS excels with multi-view inputs, its performance degrades for single-image reconstruction due to initialization ambiguity and viewpoint overfitting. We circumvent this by initializing a coarse 3D scene from the input image and progressively refining it through novel view inpainting.

Several works \cite{sabourgoli2024spotlesssplats, Pryadilshchikov2024T3DGSRT} have investigated 3DGS training on unconstrained, in-the-wild photo collections, relaxing the conventional assumption that the input images depict a perfectly consistent static 3D world. These methods identify regions violating geometric consistency as distractors — a concept analogous to the inconsistency artifacts in our generated views. Specifically, they jointly optimize a neural network $\theta$ to predict inlier/outlier masks $\{\mathbf{M}_i\}^N_{i=1}$ for each training image through self-supervision, optimizing the splats via a masked L1 loss:

\begin{displaymath}
\underset{\mathcal{G}}{\arg \min } \sum_{n=1}^N \mathbf{M}_n^{(\mathrm{t})} \odot\left\|I_n-\hat{I}_n^{(\mathrm{t})}\right\|_1,
\end{displaymath}

where the superscript $(\mathrm{t})$ indexes the training iteration of $\mathcal{G}$ and $\odot$ denotes the Hadamard product. Building upon this framework, we adapt their distractor prediction mechanism to serve as a learnable inconsistency checker in our pipeline, enhancing its capability to identify errors and artifacts during novel view inpainting.

\subsection{Conditional Video Diffusion Model}

Diffusion models \cite{ho2020denoisingdiffusionprobabilisticmodels, song2022denoisingdiffusionimplicitmodels} are generative models that learn to progressively denoise samples from a Gaussian distribution $x_T \sim \mathcal{N}(0, I)$), demonstrating remarkable generation capabilities across various domains. They consist of two primary components, a forward process $q$ that gradually adds noise to clean data $x_0 \sim q_0(x_0)$ through discrete timesteps $t$, and a learned reverse process $p_\theta$ where a neural network $\epsilon_\theta(x_t, t)$ predicts and removes the noise at each step.

When extended to conditional generation, the model accepts additional inputs $\textbf{c}$ (e.g., text prompts or images) that modulate the data distribution as a conditional denoiser $\epsilon_\theta(x_t, t, \textbf{c})$. To enhance conditional fidelity, classifier-free guidance \cite{ho2022classifierfreediffusionguidance} is often adapted to interpolate between conditional and unconditional predictions: 

\begin{displaymath}
\tilde{\epsilon}_\theta(x_t, t, \textbf{c}) = (1 + w)\epsilon_\theta(x_t, t, \textbf{c}) - w\epsilon_\theta(x_t, t)
\end{displaymath}

For video generation, the video data $x \in \mathbb{R}^{L\times 3\times H\times W}$ is first encoded into a latent representation $z \in \mathbb{R}^{L \times C \times h \times w}$ through a frame-wise VAE encoder. Both diffusion processes operate on the complete temporal sequence in latent space, ensuring inter-frame consistency while maintaining computational efficiency. The decoded output is then reconstructed through the VAE decoder. In our framework, we strategically treat multiview renderings from smooth camera trajectories as temporal video frames, thereby enabling the video diffusion model to inpaint and refine novel views while preserving consistency.

\section{Methods}

\subsection{Initializing a Coarse Scene}
\label{sec:init}

\begin{figure}[h!]
    \centering
    \includegraphics[width=\linewidth]{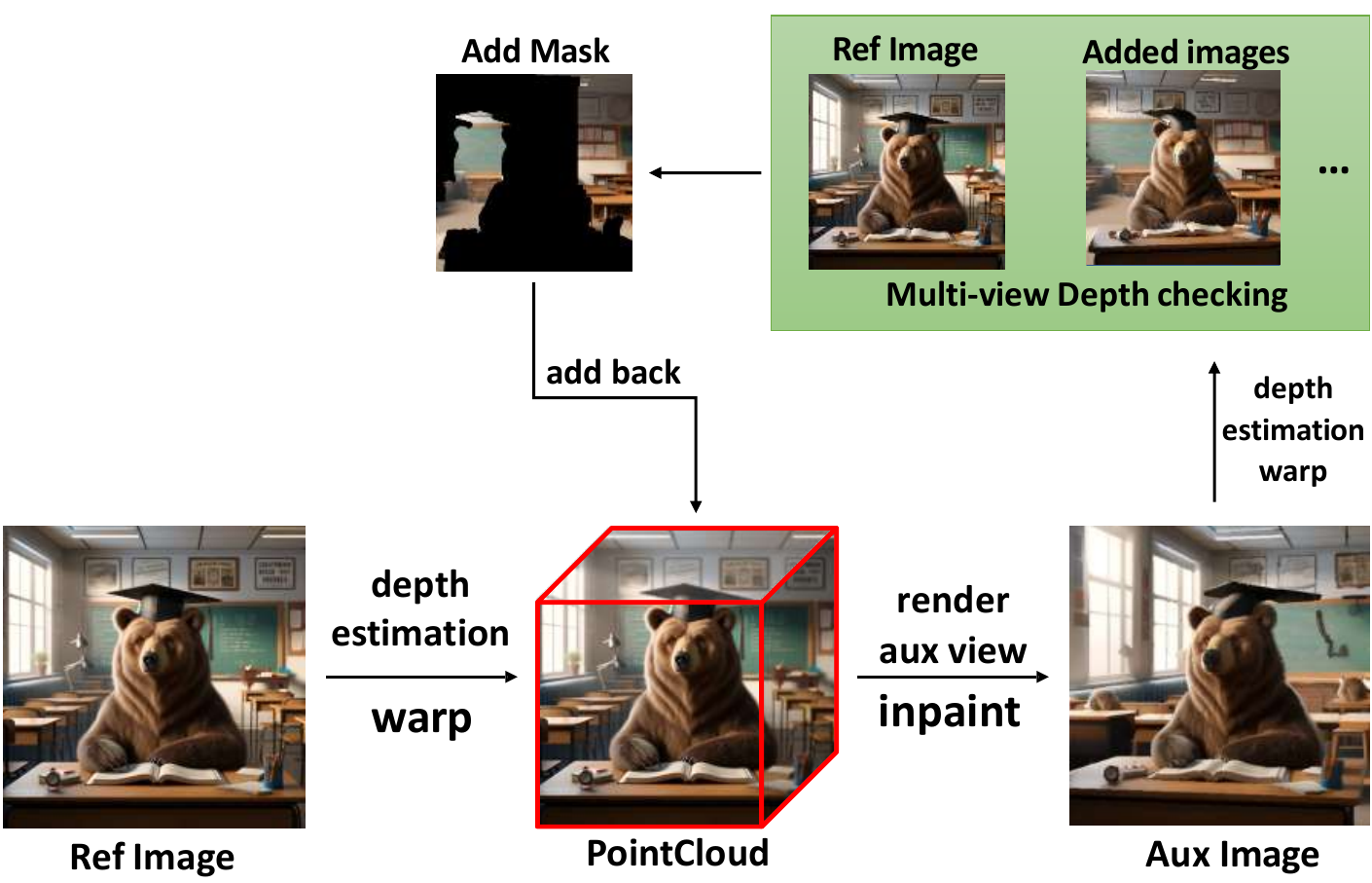}
    \Description{Our pipeline of coarse video initialization}
    \caption{Overview of initialization pipeline. We first lift the input image to a point cloud $\mathcal{P}$,  then render $\mathcal{P}$ at auxiliary viewpoints, and inpaint the occluded regions. We warp the newly inpainted image to existing views and validate geometry through depth verification. Only geometrically consistent regions (via add mask) are added to $\mathcal{P}$, yielding a coarse 3D scene $\mathcal{P}_{aux}$ that initializes both our video generation and IA-GS.}
    \label{fig:init-pipeline}
\end{figure}

As no 3D awareness is applied to the video diffusion model, the initialization quality significantly influences the final output fidelity. To generate a coarse video sequence depicting occluded regions from the single-view input image under smooth novel view trajectories, we formulate this as an inpainting problem addressed through a warp-and-inpaint paradigm, as shown in Fig. \ref{fig:init-pipeline}. Specifically, we first estimate the depth map and focal length of the reference image $I_{ref}$ using a monocular depth estimation model. This allows us to construct a point cloud $\mathcal{P}$ aligned with the predefined reference camera pose $C_{ref}$. By rendering $\mathcal{P}$ along a trajectory of $F$ novel viewpoints $\{C_i\}_{i=1}^F$, we obtain rendered images $\{I_i\}_{i=1}^F$, corresponding depth maps $\{D_i\}_{i=1}^F$, and pixel masks $\{M_i^{pix}\}_{i=1}^F$ indicating regions without points. 

However, directly using $\{M_i^{pix}\}_{i=1}^F$ for inpainting introduces perspective errors since these masks are subsets of the occluded regions of objects, which causes quality degradation in diffusion models without 3D awareness. To resolve this, we employ the occlusion volume technique \cite{shriram2024realmdreamer} to construct a volume $\mathcal{O}$ where all voxels occluded from $C_{ref}$ are assumed to contain points. Rendering $\mathcal{O}$ along $\{C_i\}_{i=1}^F$ yields occlusion depths $\{\mathcal{D}_i\}_{i=1}^F$, from which we derive the occlusion-aware inpainting masks:

\begin{equation}
    M_i^{occ} = M_i^{pix} \cup (D_i > \mathcal{D}_i)
\end{equation}

To perform a high-quality initialization for the occluded regions of the scene, we choose several auxiliary viewpoints $\{C_{aux_i}\} \subset \{C_i\}_{i=1}^F$, adapting a progressive view expansion strategy that incrementally processes auxiliary viewpoints from small to large camera offsets. At each viewpoint $aux_i$, we utilize an image diffusion model to inpaint the occlusion regions with mask $M^{occ}_{aux_i}$ and conduct depth estimation and multi-view consistency checking through depth projection. Only pixels not occluding existing points in prior views are added to the point cloud $\mathcal{P}$. 

Once the initial point cloud $\mathcal{P}_{aux}$ is filled, we render the novel view images, then inpaint them using a video diffusion model with per-frame pixel masks $\{M_i^{pix}\}_{i=1}^F$ to produce a coarse but globally consistent scene video $\{I^{init}_i\}_{i=1}^F$. Crucially, we maintain separate treatment for reference-derived points versus newly added points, generating view-specific refinement masks  $\{M_i^{refine}\}_{i=1}^F$ for subsequent processing. 

\subsection{Inconsistency Aware Gaussian Splats}
\label{sec:iags}

We then use the coarse point cloud $\mathcal{P}_{aux}$ to initialize our Inconsistency Aware Gaussian Splats, which develop an inconsistency detection framework tailored for 3D scene inpainting tasks. We follow existing methods to use a self-supervised MLP predictor $\phi$ trained with dynamically generated masks derived from the rendering loss  $R_i^{(\mathrm{t})}$ at every training iteration $(\mathrm{t})$ of Gaussian Splats $\mathcal{G}$. The generated mask is computed as 

\begin{equation}
\label{eq:res_mask}
M_i^{(\mathrm{t})}=\mathbf{1}\left\{\left(\mathbf{1}\left\{R_i^{(\mathrm{t})}>\rho\right\} \circledast \mathbf{B}\right)>0.5\right\}, P\left(R_i^{(\mathrm{t})}>\rho\right)=\tau
\end{equation}

where $\rho$ denotes the generalized median threshold (percentile $\tau$) and $\mathbf{B}$ represents a 3 $\times$ 3 box filter that performs a morphological dilation via convolution ($\circledast$). The predictor $\phi$, implemented as per-pixel 1 $\times$ 1 convolutions $\mathcal{H}$, is optimized through a bounded supervision loss that incorporates upper and lower bounds ($\mathbf{U}^{(\mathrm{t})}$,  $\mathbf{L}^{(\mathrm{t})}$) computed from Eq.(\ref{eq:res_mask}) with $\tau_{\text{low}}$ and $\tau_{\text{high}}$ respectively, and the loss $\mathcal{L}_{sup}(\phi^{(\mathrm{t})})$ is computed as:

\begin{equation}
\begin{aligned}
\mathcal{L}_{sup}\left(\phi^{(\mathrm{t})}\right) & =\max \left(\mathbf{U}^{(\mathrm{t})}-\mathcal{H}\left(\mathbf{F} ; \phi^{(\mathrm{t})}\right), 0\right) \\
& +\max \left(\mathcal{H}\left(\mathbf{F} ; \phi^{\mathrm{t})}\right)-\mathbf{L}^{(\mathrm{t})}, 0\right)
\end{aligned}
\end{equation}

Our approach introduces two key adaptations for 3D generation scenarios: 

\subsubsection{Prior-aware confidence weighting}
Our framework incorporates consistency confidence priors, where geometrically warped points (e.g., from the reference view) are inherently more reliable than diffusion-generated content in occluded regions.  We materialize this prior through two mechanisms: the preservation loss

\begin{equation}
\mathcal{L}_{prior}(\phi^{(\mathrm{t})}) = \max \left( M^p - \mathcal{H}(\mathbf{F}; \phi^{(\mathrm{t})}), 0 \right)
\end{equation}

with $M^p = \neg M^{occ}$ protects high-confidence areas, while the discard loss

\begin{equation}
\mathcal{L}_{discard}(\phi^{(\mathrm{t})}) = - \left( M^p \odot (1- \mathcal{H}(\mathbf{F}; \phi^{(\mathrm{t})}))  \odot R_i^{(\mathrm{t})} \right)
\end{equation}

actively removes the most inconsistent content. These are combined as our final loss $\mathcal{L}_{mask}(\phi^{(\mathrm{t})})$:

\begin{equation}
\begin{aligned}
\mathcal{L}_{mask} = & \mathcal{L}_{sup}\left(\phi^{(\mathrm{t})}\right) + \lambda_{prior}\mathcal{L}_{prior}(\phi^{(\mathrm{t})}) \\ 
& + \lambda_{discard}\mathcal{L}_{discard}(\phi^{(\mathrm{t})})
\end{aligned}
\end{equation}

\subsubsection{Dynamic inconsistency handling} 
Unlike conventional 3DGS methods processing fixed image sets, our MLP predictor undergoes periodic resetting after each video refinement process to escape local optima from previous iterations.  Concurrently, we progressively tighten the supervision bounds to enforce shrinking inconsistency regions as the video converges toward higher consistency. This dual strategy of predictor resetting and bound annealing ensures adaptive re-localization of emerging artifacts.

\begin{figure}[h]
    \centering
    \includegraphics[width=\linewidth]{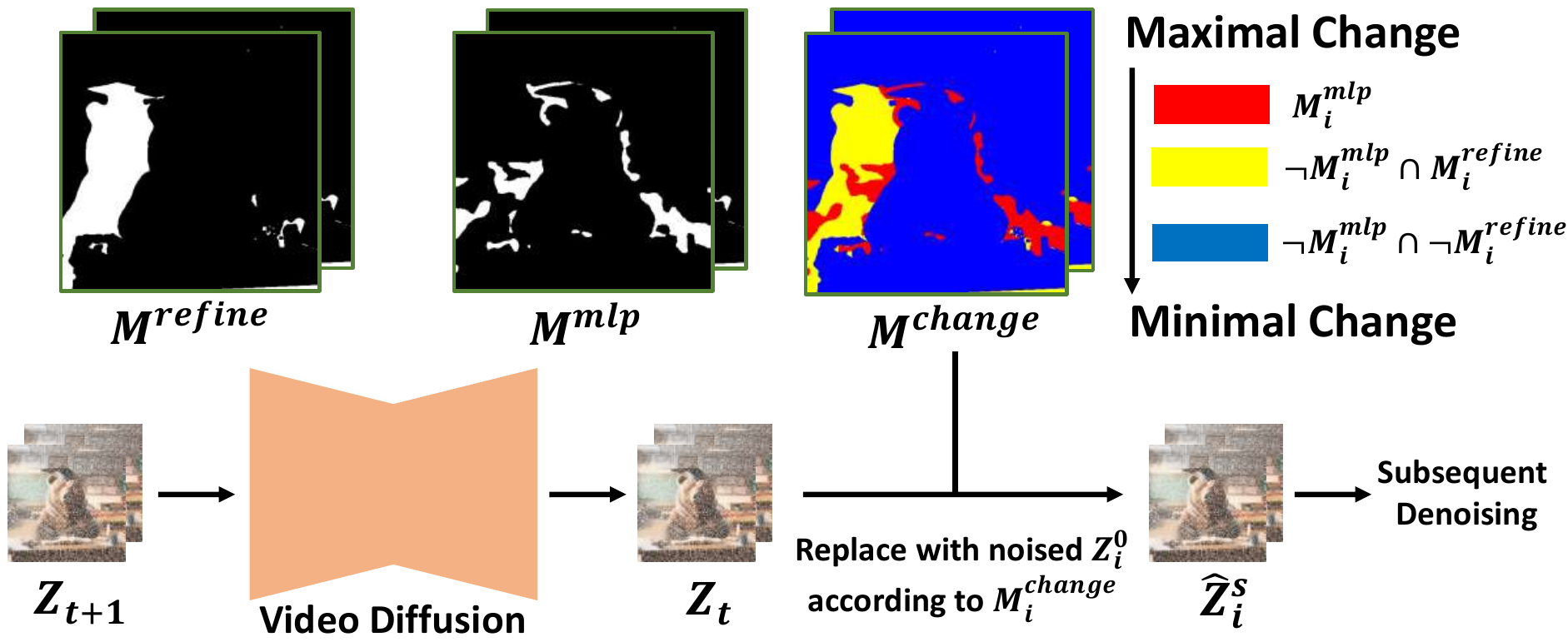}
    \Description{Our pipeline of  video refine}
    \caption{Overview of our video refinement method. We compute progressive change maps from inconsistency-aware masks and refinement masks to guide inpainting. We progressively integrate reliable multiview evidence for higher-quality output. }
    \label{fig:refine}
\end{figure}

\subsection{Video and Scene Refinement}
\label{sec:refine}

Our initial coarse scene suffers from both geometric and semantic errors (e.g., lighting discontinuities) due to per-view initialization. While the inconsistency-aware mask $M^{mlp}= \text{binary}(1- \mathcal{H}(\mathbf{F}; \phi^{(\mathrm{t})}))$ captures geometric errors, semantic correction relies on video diffusion. We render all views $\{ \hat{I}_i^{(\mathrm{t})} \}_{i=1}^F$ with the current 3DGS model, encode to latents $z = \{\mathcal{E}(I_i^{(\mathrm{t})})\}$, then noisify to $z_t$ at random timestep $t$ from the diffusion model's noise schedule.

We observe that directly applying binary masks $M^{mlp} \ \cup \ M^{refine}$ to video refinement causes global inconsistencies due to the diffusion model's lack of 3D awareness, which generates distorted content in occluded areas that contradicts geometrically visible regions in other viewpoints. To mitigate this, we first provide richer contextual clues for occluded region by replacing binary masks with progressive change maps $M^{change}$ , as shown in Fig. \ref{fig:refine}. We apply maximal change for geometrically inconsistent regions ($M^{mlp}$), mediate change for consistent generated regions ($\neg M^{mlp} \cap M^{refine}$), and minimal change for validated visible areas ($\neg M^{mlp} \cap \neg M^{refine}$). This is achieved by adjusting the inference chain length \cite{levin2023differential} - longer chains induce more modifications, allowing the diffusion model to assimilate multiview evidence from reliable regions progressively. Formally, for pixels with change weight $\text{w}^{change}$, we substitute the predicted latent $z_t$ with the noised original latent $z_0$ on the corresponding coordinate when the current timestep $t$ satisfies $\frac{T-t}{T} < 1 - \text{w}^{change}$. Besides, we incorporate estimated depths on current video as an additional condition to prevent excessive change in the geometric consistent regions $\neg M^{mlp}$. Then, refined latents are obtained via:

\begin{equation}
\hat{z} = \textrm{Diffusion}(z_t,\ M^{change}, \ \text{depth},\ \text{text})
\end{equation}

The refined latents $\hat{z}$ are decoded to $\{\tilde{I}_i\} = \mathrm{D}(\hat{z})$. We reset the MLP predictor to clear accumulated inconsistency estimates and optimize the IA-GS model using a composite loss function that balances refinement with preservation:

\begin{equation}
\begin{aligned}
\mathcal{L}_{gs} & = \|( M^{mlp}\cup  M^{refine}) \odot (\tilde{I}-\hat{I})\|_2^2 \\
& + \|\neg( M^{mlp} \cup  M^{refine}) \odot (I-I^{init})\|_1 \\
& - \textbf{Pearson}(D, \hat{D})
\end{aligned}
\end{equation}

where we conduct L2-driven quality improvement in refined regions, L1-based preservation of validated content, and depth correlation enforcement for 3D coherence. 

\begin{figure*}[h!]
    \centering
    \includegraphics[width=\linewidth]{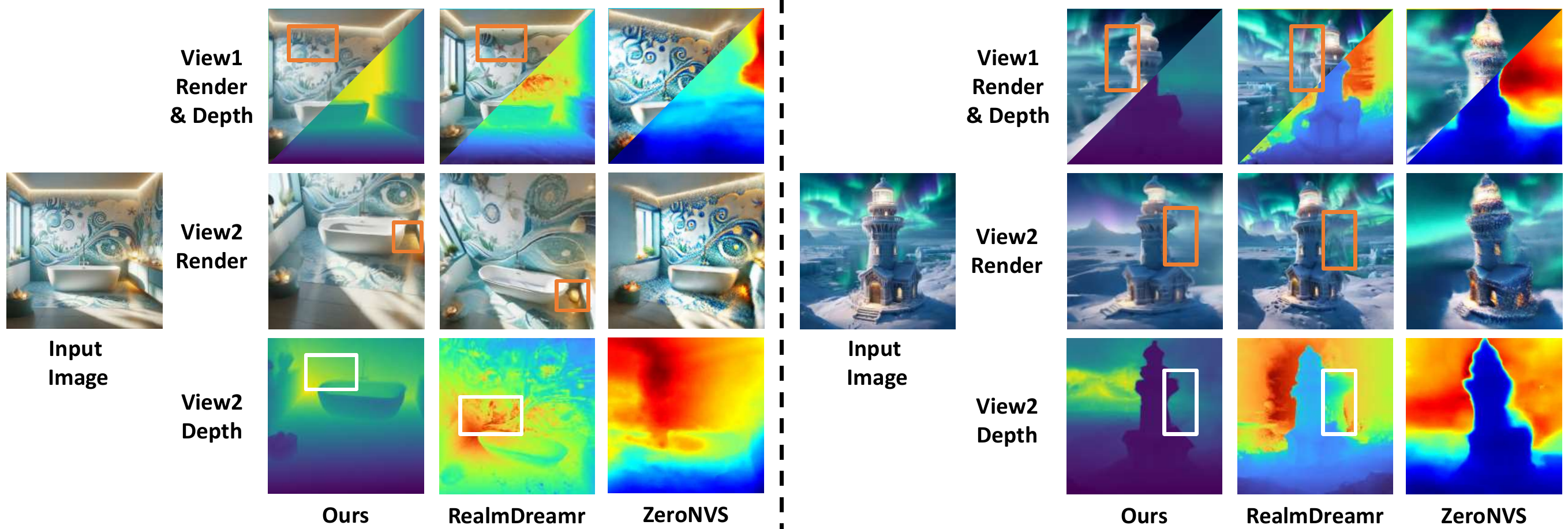}
    \Description{Qualitative Comparison}
    \caption{Qualitative Comparison. Due to our rearranged camera trajectory, the visualization is constrained to nearest-neighbor viewpoints when comparing methods.  Our method generates more plausible geometry and imagery than ZeroNVS \cite{zeronvs} and RealmDreamer \cite{shriram2024realmdreamer}, with significantly reduced computation time.}
    \label{fig:quali_cmp}
\end{figure*}

\begin{table*}[h]
  \caption{Quantitative evaluations on scene renderings of our method and the baselines.}
  \label{tab:quantitative}
  \begin{tabular}{lcccccc}
    \toprule
    Method & CLIP $\uparrow$ & Depth Pearson $\uparrow$ & LLaVA-IQA Structure $\uparrow$ & LLaVA-IQA Quality $\uparrow$ & Train Diffusion & Time $\downarrow$ \\
    \midrule
    ZeroNVS & 25.61 & 0.82 & 0.371 & 0.390 & $\checkmark$ & 1h \\
    RealmDreamer & 31.69 & 0.89 & 0.325 &  0.431 & $\times$ & 13 hours \\
    Ours-Coarse & 27.72& - & 0.659& 0.528& $\times$& 5 min\\
    Ours & 29.52 & 0.97 & 0.763 & 0.572 & $\times$ & 25 min \\
    \bottomrule
  \end{tabular}
\end{table*}

\section{Experiments}

\subsection{Experimental protocol}
\paragraph{Datasets} 

Our evaluation employs 20 generated single-view images with corresponding text prompts from our baseline \cite{shriram2024realmdreamer}. This synthetic dataset eliminates data acquisition biases and enables rigorous testing of generalization capabilities on novel scene compositions. For video diffusion processing, we rearrange camera trajectories to ensure smooth viewpoint transitions while maintaining physically plausible poses.

\paragraph{Baselines} 

We compare our method with concurrent 3D generation approaches based on image diffusion without embedded inconsistency awareness: ZeroNVS \cite{zeronvs} and RealmDreamer \cite{shriram2024realmdreamer}. ZeroNVS employs a fine-tuned multiview diffusion model to generate novel views conditioned on camera transformations, followed by NeRF optimization using Score Distillation Sampling (SDS) \cite{poole2022dreamfusiontextto3dusing2d}. RealmDreamer progressively expands a 3D Gaussian field through iterative diffusion-based inpainting, incorporating both RGB and depth supervision.

\paragraph{Metrics} 

Since ground truth data from novel viewpoints is unavailable, conventional reconstruction metrics (e.g., PSNR, LPIPS \cite{zhang2018unreasonableeffectivenessdeepfeatures}) become inapplicable for evaluation. Therefore, we adopt three complementary assessment metrics: (1) semantic alignment with text prompts measured by CLIP similarity scores \cite{radford2021learningtransferablevisualmodels}, (2) geometric consistency quantified through Pearson correlation between rendered depth and DepthAnythingV2 \cite{depth_anything_v2} predicted depth, and (3) perceptual quality evaluation via LLaVA-IQA \cite{wang2024vistadreamsamplingmultiviewconsistent}, where a vision-language model systematically assesses image structure and rendering quality through targeted questioning. This multi-faceted evaluation framework enables a comprehensive quantification of both appearance preservation and geometric fidelity in the absence of ground truth references.

\paragraph{Implementation Details} 

Our pipeline's implementation consists of the following key components: (1) The coarse stage employs PyTorch3D \cite{ravi2020accelerating3ddeeplearning} for point cloud processing, Stable Diffusion 2.0 \cite{rombach2022high} for inpainting, and DepthPro \cite{bochkovskii2024depthprosharpmonocular} for monocular depth estimation. (2) For video diffusion, we adopt AnimateDiff \cite{guo2024animatediffanimatepersonalizedtexttoimage}, which integrates Temporal Transformer modules into a Stable Diffusion 1.5 backbone. (3) Depth conditioning is achieved using Video-Depth Anything \cite{chen2025videodepthanythingconsistent} to predict per-frame depth maps. (4) Our 3D Gaussian Splatting (3DGS) implementation builds upon SpotLessSplat \cite{sabourgoli2024spotlesssplats}. Each scene undergoes 15k training iterations, with video refinement every 2k steps. During optimization, we linearly anneal the noise level $s$ from 0.6 to 0.3 and tighten the lower bound $\tau_{\text{low}}$ from 0.7 to 0.85. The entire process completes in 25 minutes on a single NVIDIA RTX 4090D (24GB VRAM).

\subsection{Qualitative Results}

Fig \ref{fig:quali_res} demonstrates GaussVideoDreamer's capability to generate high-fidelity 3D scenes with physically plausible geometry across both indoor and outdoor settings. And comparative results in Fig. \ref{fig:quali_cmp} reveal three key advantages: (1) In bathroom scene, our method produces significantly smoother surfaces and accurately reconstructs wall patterns where RealmDreamer fails; (2) For lighthouse scene, GaussVideoDreamer correctly infers occluded regions while RealmDreamer's output leads to 3DGS collapse; (3) Although ZeroNVS generates plausible low-frequency structures, it struggles with blurred geometric edges and missing background elements compared to our approach.

\subsection{Quantitative Metrics}

Our quantitative results (Table \ref{tab:quantitative}) demonstrate consistent improvements across all metrics except for an acceptable CLIP score decrease, with particularly significant gains in geometric accuracy (Depth Pearson from 0.89 to 0.97) and structural integrity (LLava-IQA Structure from 0.325 to 0.763). Our method achieves these superior reconstruction qualities while drastically reducing computational requirements from 13 hours to just 25 minutes, concurrently improving overall LLaVA-IQA Quality from 0.431 to 0.572. These comprehensive advancements validate the effectiveness of our video diffusion framework combined with inconsistency-aware Gaussian splatting for high-fidelity 3D scene reconstruction.

\begin{table}[h]
\centering
\caption{Ablation Study Results}
\label{tab:ablation}
\begin{tblr}{
  column{even} = {c},
  column{3} = {c},
  column{5} = {c},
  hline{1,5} = {-}{0.08em},
  hline{2} = {-}{0.05em},
}
Method    & CLIP  & {Depth \\Pearson} & {LLaVA-IQA \\Structure} & {LLaVA-IQA\\Quality} \\
w/o Init  &  28.82  &  0.94 & 0.282 & 0.164 \\
w/o IA-GS & 30.08 & 0.96 & 0.671 & 0.515 \\
Ours & 29.52 & 0.97 & 0.763 & 0.572                
\end{tblr}
\end{table}

\subsection{Ablations}

\begin{figure}[h]
    \centering
    \includegraphics[width=\linewidth]{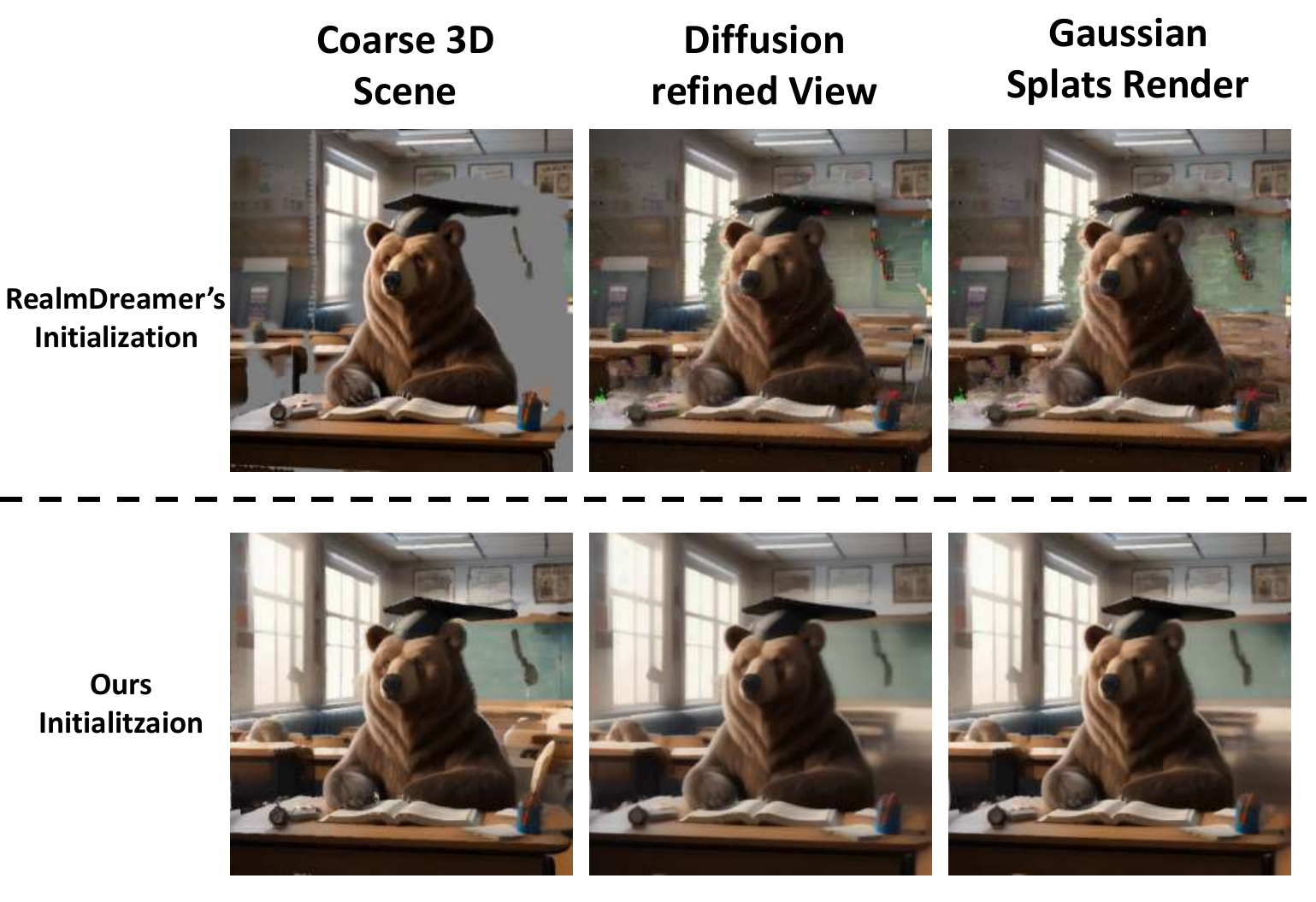}
    \Description{Ablation study on initialization}
    \caption{Compared to RealmDreamer's gray-splat initialization causing inconsistent generation and 3DGS failures, our occlusion-aware inpainting enables more stable 3D/video diffusion refinement.}
    \label{fig:init-ablation}
\end{figure}

\begin{figure}[h]
    \centering
    \includegraphics[width=\linewidth]{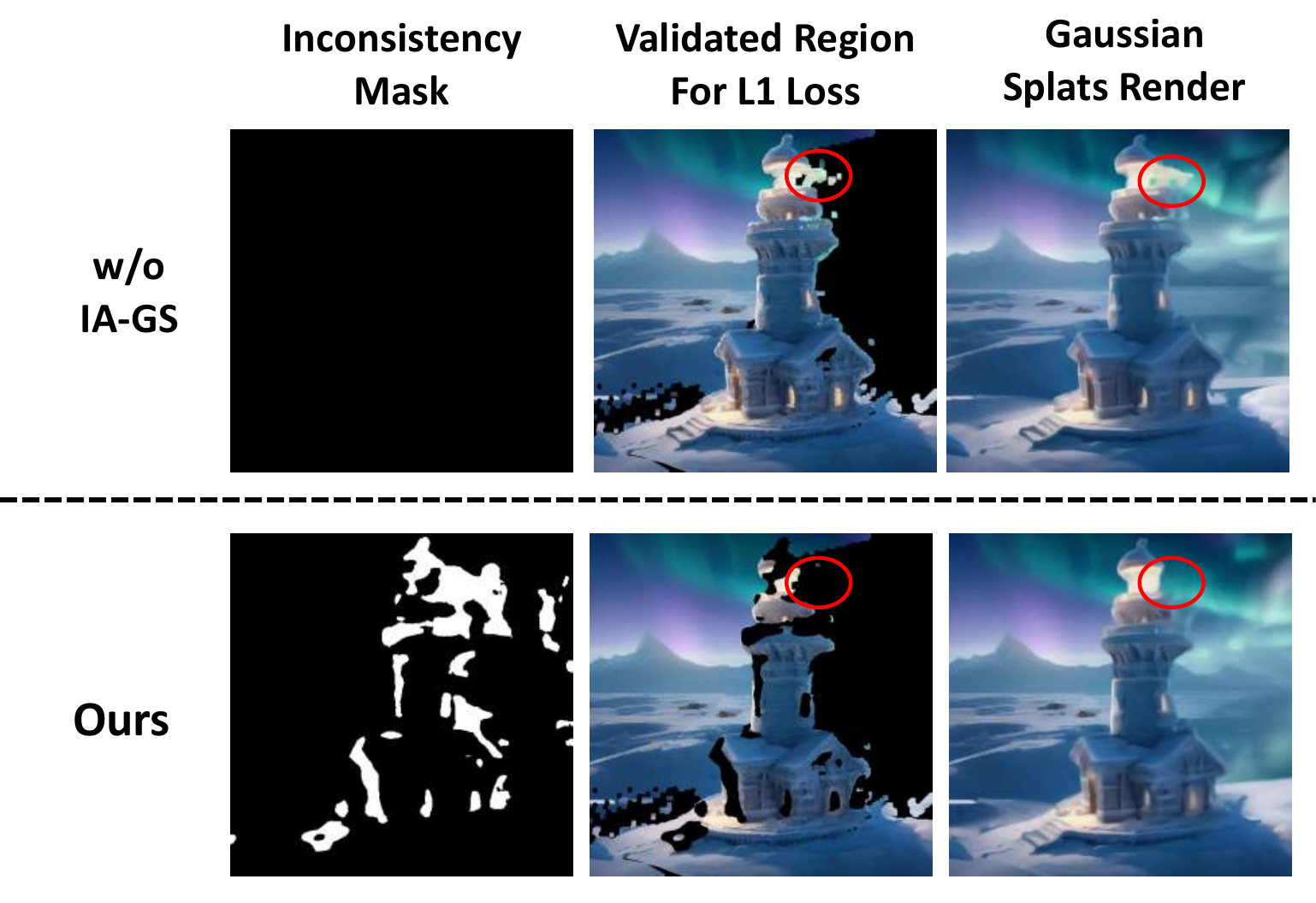}
    \Description{Ablation study on gs}
    \caption{Ablation Results on Inconsistency-Aware GS. Our IA-GS module identifies and rectifies erroneously projected regions through targeted diffusion-based erasure, significantly enhancing 3D reconstruction quality.}
    \label{fig:gs-ablation}
\end{figure}

We conduct an ablation study to validate two critical designs: our coarse video initialization strategy and  Inconsistency-Aware GS. First, when replacing our initialization with RealmDreamer's approach, which depends solely on video diffusion to hallucinate occluded regions without initialized inpainting and warping. The absence of 3D consistency constraints in the video diffusion leads to severe distortions in occluded areas, ultimately causing catastrophic failure in 3DGS optimization (Fig. \ref{fig:init-ablation}). Second, ablation of our IA-GS (reverting to vanilla 3DGS) results in persistent floating artifacts due to undetected erroneous projections from the coarse stage (Fig. \ref{fig:gs-ablation}). Quantitative results (Table \ref{tab:ablation}) reveal that our initialization improves all key metrics and IA-GS achieves superior scene integrity with a marginal CLIP score reduction. These comparisons demonstrate that both proposed components are indispensable for achieving geometrically stable and high-quality synthesis when integrating video diffusion into 3D generation pipelines.

\section{Conclusion an Limitation}

\paragraph{Conclusion}

We present GaussVideoDreamer, a novel pipeline for single-image 3D generation that harnesses video diffusion while addressing its consistency limitations.  Our framework first establishes a geometrically stable coarse reconstruction through iterative view inpainting, then introduces Inconsistency-Aware Gaussian Splatting to dynamically detect 3D errors, and utilizes video diffusion to refine the scene with its temporal-to-view coherence adaptation. Without the requirement to fine-tune the diffusion model, our method outperforms existing approaches in both quality and efficiency, as validated by multiple qualitative results and quantitative metrics.

\paragraph{Limitation} 

Our generation quality is primarily limited by the video diffusion prior (AnimateDiff), which was adapted from an image diffusion model lacking dedicated video inpainting training \cite{guo2024animatediffanimatepersonalizedtexttoimage}. Its limited video generation capability manifests as frame inconsistencies in longer sequences, progressive color shifts, and high-frequency detail degradation. Our pipeline consequently depends on high-quality initialization and constrained refinement iterations to ensure optimization stability — a limitation that could be addressed through improved video diffusion architectures in future work.

\bibliographystyle{ACM-Reference-Format}
\bibliography{main}

\end{document}